# Wrapper Feature Selection Algorithm for the Optimization of An Indicator System of Patent Value Assessment

Yihui Qiu, Chiyu Zhang*

**Abstract:** *Effective patent value assessment provides decision support for patent transection and promotes the practical application of patent technology. The limitations of previous research on patent value assessment were analyzed in this work, and a wrapper-mode feature selection algorithm that is based on classifier prediction accuracy was developed. Verification experiments on multiple UCI standard datasets indicated that the algorithm effectively reduced the size of the feature set and significantly enhanced the prediction accuracy of the classifier. When the algorithm was utilized to establish an indicator system of patent value assessment, the size of the system was reduced, and the generalization performance of the classifier was enhanced. Sequential forward selection was applied to further reduce the size of the indicator set and generate an optimal indicator system of patent value assessment.*

## I. Introduction

Effective assessment of patent value is a crucial part of patent commercialization and industrialization, and it directly relates to the revenue of developing or buying patent technology. Hence, a common challenge faced by scholars and patent inventors is efficiently and accurately evaluating patent value.

An indicator system of patent value assessment is a set of indicators that reflect the overall characteristics of a patent. To investigate the different characteristics of a patent and evaluate the patent value, this system compares the differences among patents in each component and analyzes a patent from various perspectives; it is the most widely applied method of patent value assessment.

Two international patent value indicator systems are utilized; one is the indicator system based on hypothetical methods, and it includes Georgia-Pacific factors [1] and the CHI indicator system [2]. The other is the indicator system based on machine learning, and an example is the Ocean Tomo 300® Patent Index (OT300) [3]. The "Indicator System of Patent Value Analysis" published in 2011 by the State Intellectual Property Office (SIPO) of the People's Republic of China [4] uses three level-one indicators, namely, technological, legal, and economic values, and a series of level-two indicators to provide prior knowledge for the analysis of patent value.

Many studies have identified indicators for the assessment of patent value. Schankerman [5] and Lanjouw [6] empirically and repetitively studied French and German patents and reported that technical category and nationality of the applicant(s) are the main factors that influence patent value. Bessen [7] employed "renewal decision of a patent" to assess patent value. Hall et al. selected R&D investment and success to estimate patent value. Feng et al. used assessment methods based on the analysis of citation networks [9–10]. They assessed patent value by examining direct citations and latent citation networks. Yi Zhang et al. constructed an entropy-based indicator system, used Shannon's entropy theory to weigh indicators quantitatively, and identified the four primary factors that make up the indicator system; these four are number of legal transactions, number of claims, time gap, and number of citations. An increasing number of scholars have investigated the use of machine learning algorithms in the assessment of patent value and establishment of indicator systems. Ercan and Kayakutlu [12] applied support vector machine (SVM) to analyze indicators. Wu et al. [13] proposed an automatic patent quality analysis and classification system that combines self-organizing maps, kernel principal component analysis, and SVM and concluded that assignee, patent forward citation, and claims are critical factors in patent value analysis. Ysao et al. [14] used a patent classification model based on an artificial immune system (AIS) to investigate legal status, International Patent Classification-Current, count of family members, and 19 other indicators. Clustering algorithms have also been widely applied to analyze the characteristics of patents [15–16]. However, these methods cannot provide a transparent and efficient assessment rule, and a consistent indicator system of patent value assessment remains lacking.

In this study, we developed a wrapper-mode feature selection algorithm that is based on the prediction accuracy of classification models and utilized this algorithm to establish an indicator system of patent value assessment. This feature selection algorithm efficiently solved the contradiction between dimensionality reduction and generalization errors of the learning model. The results of experiments on UCI standard datasets indicated that the algorithm could be applied to different classifier algorithms; it effectively reduced the size of the feature set and enhanced the generalization performance of the classifiers. We used this feature selection algorithm and the classification and regression tree (CART) [17] algorithm to primarily select indicators and obtain a primarily selected feature set. Based on the primarily selected feature set, an optimal feature set was constructed by using sequential forward selection (SFS) [20–21] with CART, SVM [18], and *k*-nearest neighbor (*k*-NN) [19]. This optimal feature set was the final indicator system of patent value assessment.

## II. Feature Selection Algorithm Based on the Prediction Accuracy of Classification Models

Feature selection is divided into three modes[22]. The first is filter mode, which independently selects data features and is irrelevant to the subsequent learning algorithm. The second is wrapper mode, a type of feature selection that uses the performance of models as the evaluation criteria for feature selection. The third is embedding mode, in which the process of feature selection is integrated into the process of learning model training. As indicated by the performance of final learning models, the wrapper mode is generally better than the filter mode because the wrapper mode directly optimizes the learning model. The wrapper mode is also more flexible and explicit than the embedding mode.

We developed a wrapper-mode feature selection algorithm based on the prediction accuracy of classification models and employed this algorithm in the selection of patent value assessment indicators. The basic idea of this algorithm is that the importance of a feature is defined by the influence degree of the accuracy of classifiers. A high degree of impact indicates that the feature is highly important in the prediction task, and a low degree of impact indicates that the feature is only slightly relevant to the prediction task (i.e., not highly critical). Feature selection was implemented based on this idea. First, the "influence coefficient" of a feature was defined by the degree of impact of classifier accuracy after removing this feature. Second, the "lifting coefficient" of a feature was defined by the lifting degree of classifier accuracy after adding this feature. To consider the two coefficients simultaneously, we determined the factors that are crucial to the prediction task.

The influence coefficient is defined in Eq. (1).

$$\overline{A_i} - \overline{A_0} = I_i , \qquad (1)$$

where $\overline{A_0}$ denotes the mean accuracy of the initial model, $\overline{A_i}$ denotes the mean accuracy of a model after removing the *i*th feature, and a high $|I_i|$ indicates that the removed feature is highly important.

The initial classifier was trained by the original feature set, and 5-fold cross-validation was applied to calculate the prediction accuracy of classifiers. The training process was executed repeatedly for *n* times, and *n* accuracies were averaged to acquire the mean of accuracies, i.e., $\overline{A_0}$, which is the mean accuracy of the initial model without feature selection, as presented in Eq. (2).

$$\overline{A_0} = \frac{\sum_{j=1}^{n} A_{0j}}{n} , \qquad (2)$$

where $A_{0j}$ is the prediction accuracy of the *j*th execution and *n* is the time of executions.

The *i*th feature was removed from the original feature set to determine the importance of this feature. A classifier was trained using the dataset without the *i*th feature, and 5-fold cross-validation was implemented to calculate the prediction accuracy of the classifier. The training process was repeatedly executed *n* times for every removed feature, and *n* accuracies were averaged to obtain the mean prediction accuracy of the removed *i*th feature $\overline{A_i}$. The results were substituted into Eq. (1) to obtain the influence coefficient of the *i*th feature. The influence coefficients of all features were generated to remove each feature of the original feature set in an orderly manner. The features were ranked depending on their importance in accordance with the value of the influence coefficients.

If feature importance is only defined by the degree of impact of classifier accuracy after removing a feature, then overfitting of models and bias will arise. Therefore, the "lifting coefficient" was introduced to dialectically measure the lifting degree of prediction accuracy of each feature and accurately determine the importance of each feature. The "lifting coefficient" is defined in Eq. (3).

$$\overline{B_i} - \overline{B_{i-1}} = P_i , \qquad (3)$$

where $\overline{B_{i-1}}$ is the mean of accuracies before adding the *i*th feature, $\overline{B_i}$ denotes the mean of accuracies after adding the *i*th feature, and $P_i$ is the lifting coefficient of the *i*th feature. When $P_i > 0$, the added feature improves the accuracy of the prediction model. A high $P_i$ indicates that the lifting degree is high, and the *i*th feature is important. When $P_i \leqslant 0$, the feature is irrelevant to the prediction task.

According to the ranking of the influence coefficient of features, classifiers were built with the sequential forward recursive method. Features were sequentially added to the learning model, and only one feature was added in each recursion. A new classifier with the newly added $i$th feature was built, and the accuracy of the classifier after adding the $i$th feature was calculated using 5-fold cross-validation. Then, the training process was executed repeatedly for $n$ times, and the $n$ accuracies were averaged to obtain the mean of accuracies of the classifier with the $i$th feature, i.e., $\overline{B_i}$. The result was substituted into Eq. (3) to obtain the lifting coefficient of the $i$th feature. To rank the features in a descending order on the basis of the lifting coefficients, the features whose lifting coefficients are greater than 0 were selected. The final feature set was determined to be the optimal feature subset by using the feature selection algorithm.

## III. Simulation Experiments

The feature selection algorithm presented above adequately considers the diversity of classifier algorithms and the consistency of classification tasks and efficiently solves the contradiction between dimensionality reduction and generalization errors of the classifier. Eight UCI standard datasets [23] were used to verify the validity of the feature selection algorithm. Verification experiments were conducted on the eight UCI standard datasets, all of which contained classification tasks. The characteristics of the datasets are shown in Table 1.

Table 1. Description of experiment datasets

| Name of Dataset | Number of Features | Number of Samples | Number of Class |
|---|---|---|---|
| Zoo | 16 | 101 | 7 |
| Wine | 13 | 178 | 3 |
| Sonar | 60 | 208 | 2 |
| Waveform | 21 | 5000 | 2 |
| Ionosphere | 34 | 351 | 2 |
| Soybean | 35 | 47 | 4 |
| Segmentation | 19 | 210 | 7 |
| Hepatitis | 19 | 155 | 2 |

The eight datasets were processed by the feature selection algorithm proposed in Section II. To calculate stable and unbiased coefficients, 5-fold cross-validation was implemented to estimate the prediction accuracy of each model, and each feature set was repeatedly trained 100 times. The experiments utilized CART and $k$-NN as the basic classifiers for feature selection to verify the universality of the feature selection algorithm in different machine learning algorithms. The experiment results are shown in Table 2.

Table 2. Comparison of prediction accuracy before and after feature selection

| Name of Dataset | Feature Size Before Selection | Accuracy (%) | | | | | |
|---|---|---|---|---|---|---|---|
| | | CART | | | $k$-NN | | |
| | | Before | After | Feature Size After Selection | Before | After | Feature Size After Selection |
| Zoo | 16 | 88.82±0.94 | 91.26±1.62 | 11 | 95.16±1.34 | 96.93±0.40 | 10 |
| Wine | 13 | 92.57±1.89 | 94.38±1.44 | 8 | 95.31±0.54 | 97.12±0.71 | 8 |
| Sonar | 60 | 73.73±2.80 | 76.62±2.35 | 38 | 85.97±1.56 | 88.05±1.16 | 35 |
| Waveform | 21 | 76.23±0.46 | 76.31±0.43 | 13 | 76.24±0.26 | 78.81±0.29 | 14 |
| Ionosphere | 34 | 88.67±1.29 | 90.06±1.24 | 27 | 86.43±0.91 | 89.19±0.80 | 17 |
| Soybean | 35 | 98.85±1.90 | 100.00±0.00 | 30 | 99.87±0.37 | 100.00±0.00 | 28 |
| Segmentation | 19 | 86.04±1.93 | 89.21±1.86 | 18 | 84.61±0.98 | 89.68±1.13 | 8 |
| Hepatitis | 19 | 79.26±2.25 | 80.41±2.40 | 16 | 84.10±1.62 | 85.52±2.37 | 11 |

According to the experiment results in Table 2, the feature selection algorithm in this work not only reduced the feature sizes of the eight datasets but also effectively improved the prediction accuracy of the classifiers. By using different classification algorithms, this feature selection process presents broad applicability in different classifiers and enhances the generalization errors of classifiers. Next, we took a set of experiments with the most significant improvement in accuracy as an example to explain the processes of this feature selection algorithm. We selected the experiments on the segmentation dataset with the $k$-NN clustering algorithm as the basic classifier. According to the feature selection proposed in Section II, the feature selection processes of the segmentation dataset were as follows:

Input: Feature set $C = \{C_1, C_2 \dots C_l\}$, classification label $L$, $k$-NN clustering algorithm;

Step 1:

1.1 Use feature set $C$ and label $L$ to build the $k$-NN model, and acquire the mean of prediction accuracies $\overline{A_0}$ and standard deviation $\sigma_0$;

1.2 For each feature $C_i$

1.3 Use feature subset $Y$ without feature $C_i$ and label $L$ to build the $k$-NN model and the mean of prediction accuracies $\overline{A_i}$;

1.4 $\overline{A_i} - \overline{A_0} = I_i$;

1.5 End for

1.6 Remove the features where $I_i > \sigma_0$, ascending sort feature set $C$ depended on $I_i$, and acquire the feature set $C' = \{C'_1, C'_2 \ldots C'_i\}$.

Step 2:

2.1 For each feature $C'_i$,

2.2 Add the $i$th feature $C'_i$ into feature set $T_{i-1}$, then $T_i = T_{i-1} + C'_i$,

Use $T_i$ and $L$ to build $k$-NN model and the mean of prediction accuracy $\overline{B_i}$.

2.3 $\overline{B_i} - \overline{B_{i-1}} = P_i$;

2.4 End for

2.5 Select feature $C'_i$ where $P_i > 0$ and acquire the feature set $X$.

Output: optimal feature selection

$X = \{X_1, X_2 \ldots X_i\}$, $X \subseteq C$.

To apply the $k$-NN algorithm in building a model with the segmentation dataset, prediction accuracy was calculated by 5-fold cross-validation, and execution was repeated 100 times to obtain the mean and standard deviation of the classifier. $\overline{A_0}$ was 84.6191%, and $\sigma_0$ was 0.9796%. $Y_i$ is the feature subset after removing the $i$th feature $C_i$. The new feature set $C'$ in step 1.6 is shown in Table 3, and the features were selected by the influence coefficient. $T_i$ is the feature subset before adding the $i$th $C'_i$, $T_{i-1}$ is the feature subset after adding the $i$th $C'_i$, and $\overline{B_i}$ is the mean of prediction accuracies after addition. Step 2.3 calculates the lifting coefficient $P_i$. Then, we selected the features whose $P_i$ is greater than 0 and constituted a feature subset $X$. The experimental results are summarized in Table 4. The optimal feature subset comprised the eight features shown in Table 4, and a line chart (Graph 1) was drawn to display the variation tendency of accuracies.

Table 3. Feature sorting and feature selection based on the influence coefficient

| Ranking | Name of Feature | Influence Coefficient | Remain? | Ranking | Name of Feature | Influence Coefficient | Remain? |
|---|---|---|---|---|---|---|---|
| 1 | region-centroid-col | −7.0286 | Y | 11 | region-pixel-count | 0.1571 | Y |
| 2 | exred-mean | −0.5238 | Y | 12 | rawred-mean | 0.1905 | Y |
| 3 | saturatoin-mean | −0.3857 | Y | 13 | hedge-sd | 0.2048 | Y |
| 4 | hue-mean | −0.1095 | Y | 14 | short-line-density-2 | 0.5143 | Y |
| 5 | vegde-sd | −0.0524 | Y | 15 | short-line-density-5 | 0.6190 | Y |
| 6 | exgreen-mean | 0.0381 | Y | 16 | exblue-mean | 0.7714 | Y |
| 7 | intensity-mean | 0.0476 | Y | 17 | vedge-mean | 0.9238 | Y |
| 8 | rawgreen-mean | 0.0619 | Y | 18 | region-centroid-col | 0.0103 | N |
| 9 | value-mean | 0.0667 | Y | 19 | hedge-mean | 0.0106 | N |
| 10 | rawblue-mean | 0.0762 | Y | | | | |

Table 4. Forward recursion and feature selection

| Number of Feature | Name of Newly Added Feature | Prediction Accuracy (％) | Lifting Coefficient (％) | Remain? | Number of Feature | Name of Newly Added Feature | Prediction Accuracy (％) | Lifting Coefficient (％) | Remain? |
|---|---|---|---|---|---|---|---|---|---|
| 1 | region-centroid-col | 42.6238 | 42.6238 | Y | 10 | rawblue-mean | 89.2905 | 0.0429 | Y |
| 2 | exred-mean | 75.6048 | 32.9810 | Y | 11 | region-pixel-count | 89.2190 | −0.0714 | N |
| 3 | saturatoin-mean | 83.1238 | 7.5190 | Y | 12 | rawred-mean | 89.0143 | −0.2048 | N |
| 4 | hue-mean | 85.4952 | 2.3714 | Y | 13 | hedge-sd | 89.7905 | 0.7762 | Y |
| 5 | vegde-sd | 85.4905 | −0.0048 | N | 14 | short-line-density-2 | 88.9476 | −0.8429 | N |
| 6 | exgreen-mean | 88.2619 | 2.7714 | Y | 15 | short-line-density-5 | 87.4571 | −1.4905 | N |
| 7 | intensity-mean | 89.5952 | 1.3333 | Y | 16 | exblue-mean | 87.2333 | −0.2238 | N |
| 8 | rawgreen-mean | 89.5333 | −0.0619 | N | 17 | vedge-mean | 85.5190 | −1.7143 | N |
| 9 | value-mean | 89.2476 | −0.2857 | N | | | | | |

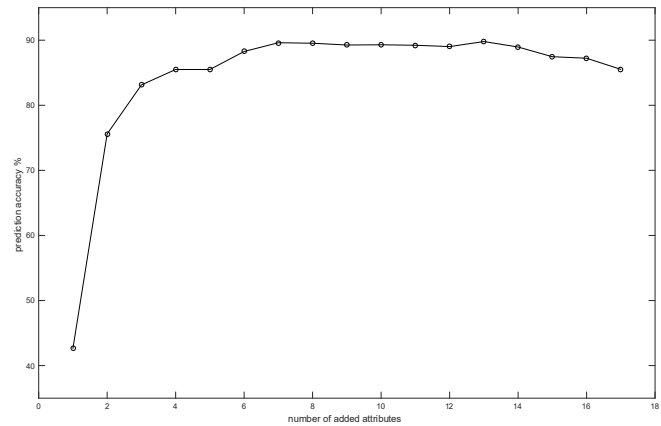

Graph 1. Variation curve of forward recursion

As shown in Table 4 and Graph 1, prediction accuracy increased with the increase in the number of features. Accuracy was at the highest when the number of features reached a certain threshold, but accuracy stabilized or even decreased gradually when the number of features exceeded the threshold. Notably, few features started to contain insufficient information to ensure that the classifier performed effectively. When the number of features exceeded the threshold, many features could not provide information to enhance classifier performance and were even mingled with redundancy and invalid information, thus affecting the generalization performance. Thus, feature selection is important, and the optimal feature subset selected for classification task can help improve classifier performance.

## IV. Establishment of an Indicator System of Patent Value Assessment

We used CART decision tree as the basic classifier and applied the feature selection algorithm proposed in Section II to primarily select indicators of patent value assessment. By using the SFS algorithm to optimally select indicators based on the feature subset from primary selection, we constructed the final indicator system of patent value assessment and analyzed it empirically.

### IV.1 Primary Selection of Patent Value Assessment Indicators Based on CART Algorithm

#### IV.1.1 CART Decision Tree

A decision tree is a hierarchy consisting of nodes and a directed edge, which includes three types of nodes, namely, root, internal, and leaf, and a clear set of decision rules, which is the path formed from the root node to leaf node [24]. Current mainstream decision tree algorithms include ID3, C4.5, SLIQ, and CART [26]. CART decision tree was proposed in 1984 by Breiman [30]. This algorithm generates a binary splitting decision tree with the Gini index as the splitting criteria, and it can deal with numeric data with high skewness or polymorphism and sequential or disordered categorical data [25]. CART decision tree possesses enhanced prediction accuracy in classification and regression tasks [26]. CART decision tree is pruned by cross validation, and the final optimal binary decision tree possesses appropriate complexity and error rate. Therefore, due to the white-box structure of the decision tree and the advantages of CART decision tree, we employed the CART algorithm as the basic classifier to establish the indicator system of patent value assessment.

#### IV.1.2 Experiment

The data in this section were derived from a professional patent information database, i.e., Technical Innovative Intelligence Platform of Incoshare [31]. Seventeen features were selected as the original dataset based on prior knowledge of "Indicator System of Patent Value Analysis," and the list of features is shown in Table 5. The value class is the classification label, which has one to nine grade levels where a high number represents a high value. By analyzing and understanding the original dataset, we combined the original one to four grade levels into one class called "class1" and each level of five to nine sets into one class. The five classes were called "class2," "class3," "class4," "class5," and "class6". The distribution of samples is presented in Table 6.

Table 5. List of features

| Number of Claims | Current Legal Statutes | Patent Type | Patent Validity | Documentation Code |
|---|---|---|---|---|
| National industrial classification | Count of simple family members | Count of extended family members | Count of family countries | Count of cited documents |
| Count of family cited documents | Count of citing times | Count of family citing times | Number of cited applicants | Count of citing applicants |
| Count of cited family applicants | Count of citing family applicants | Value class | | |

Table 6. Distribution of samples

| Class1 | Class2 | Class3 | Class4 | Class5 | Class6 | total |
|---|---|---|---|---|---|---|
| 292 | 242 | 231 | 223 | 341 | 261 | 1590 |

By using the feature selection algorithm proposed in Section II and CART as the basic classifier, we primarily selected the indicators and listed them in Table 7. A comparison of the prediction accuracies of the decision tree is shown in Table 8.

Table 7. Indicators after primacy feature selection

| Order Number | Name of Feature | Order Number | Name of Feature |
|---|---|---|---|
| a | Number of claims | b | Patent type |
| c | Documentation code | d | Count of simple family members |
| e | Count of extended family members | f | Count of cited documents |
| g | Count of family citing times | | |

Table 8. Comparison of prediction accuracy before and after primary selection

| Prediction Accuracy (%) | |
|---|---|
| Before | After |
| 53.9538±0.2100 | 90.3623±0.4714 |

## IV.2 Optimization of the Indicators of Patent Value Assessment

To improve the efficiency and performance of the classifier of patent value assessment, the SFS algorithm was used to find a small feature subset based on the indicator set after primary selection. After ensuring that the final optimal feature subset is accurate and unbiased, we used three classification algorithms to build models and generate the optimal indicator system of patent value assessment.

SFS [20–21] is a wrapper-mode feature selection algorithm that starts with an empty predictor feature set $X_0$ and a full feature set as candidate feature set $C$, i.e., $C = \{C_1, C_2 \ldots C_m\}$, where $m$ is the number of features of $C$. For the first step, each feature $C_i$ of $C$ is built as a classifier with an object function, and the accuracy of classifier $A_{1i}$ is calculated. A feature is added to $X_0$ to create a new predictor feature set $X_1$ and removed from $C$ if it gives the highest $A_{1i}$. The prediction accuracy of $X_1$ is $B_1$. For the $j$th step, the remaining features of $C$ are added individually to $X_{j-1}$, and the new feature subset is evaluated. The feature is appended to $X_{j-1}$ to create $X_j$ and removed from $C$ if it gives the highest $A_{ji}$. The prediction accuracy of $X_j$ is $B_j$. This process is repeatedly executed until $C$ becomes an empty feature set. To achieve the optimal feature subset, the prediction accuracies and sizes of $X_i$ were compared, where $i=1,2\ldots m$.

We applied CART, SVM, and $k$-NN algorithms to build models and optimize the indicator system of patent value assessment. According to the results of the parameter optimization experiments, the parameters of leading models were set as follows: the splitting criterion of CART was the Gini index; the kernel function of SVM was a quadratic kernel, where $C$ equals 1; and the clustering algorithm was the weighted $k$-NN algorithm in which the number of neighbors was 6, the distance metric was city block, and the distance weight was squared inverse.

Each feature combination was trained, and three classifiers were generated: CART, SVM, and $k$-NN. The prediction accuracies of each classifier included the mean of 100 5-fold cross-validation accuracies. In this manner, the prediction accuracies of different combinations and the optimal feature subset with the highest prediction accuracy were obtained. In Table 9, the letters of feature combination correspond to the order number in Table 7.

Table 9. Feature set reduction with SFS

| Number of Added Features | Prediction Accuracy | | | | | |
|---|---|---|---|---|---|---|
| | CART | | SVM | | *k*-NN | |
| | Highest% | Feature Combination | Highest% | Feature Combination | Highest% | Feature Combination |
| 1 | 62.4747 | c | 62.5834 | c | 37.3434 | c |
| 2 | 77.7398 | ac | 78.7655 | ac | 71.5547 | ac |
| 3 | 86.8208 | acg | 86.4395 | acg | 82.2075 | acg |
| 4 | 89.4261 | acdg | 88.7353 | acdg | 85.6277 | acdg |
| 5 | 90.7588 | abcdg | 90.6196 | abcdg | 86.8478 | abcdg |
| 6 | 90.5744 | abcdfg | 90.4623 | abcdfg | 86.3082 | abcdfg |
| 7 | 90.4142 | abcdefg | 90.1811 | abcdefg | 85.2365 | abcdefg |

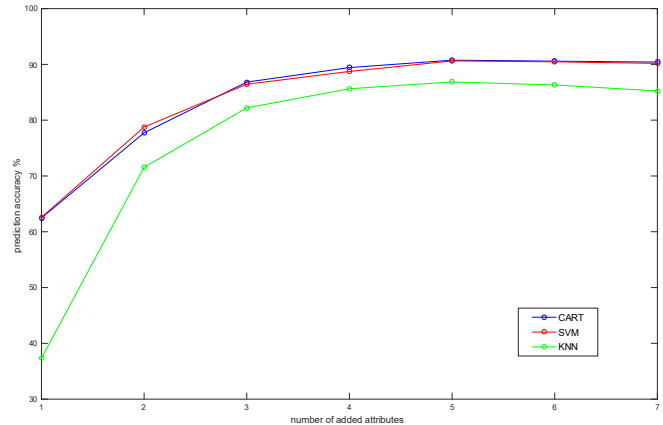

Graph 2. Process of SFS

According to the feature combinations in Tables 9 and 7, the best combinations of CART, SVM, and *k*-NN are all feature subsets that have the number of claims, patent type, documentation code, count of simple family members, and count of family citing times.

**4.3 Establishment of an Indicator System of Patent Value Assessment**

As revealed by the experiments, the best prediction accuracies of CART, SVM, and *k*-NN were obtained when the size feature subset was five, and their best feature subsets were identical. Hence, we considered these five factors the most important factors for patent value assessment. The optimal indicator subset is the subset that includes the number of claims, patent type, documentation code, count of simple family members, and count of family citing times, and this optimal feature subset is our indicator system of patent value assessment.

The indicator system of patent value assessment was then analyzed briefly.

"Number of claims" is the sum of the number of independent and dependent claims. Patent claim defines the legal protection areas of a patent. A high number of patent claims indicates rigorous limited relationships, high legal value, and small likelihood of being completely invalidated in litigation. In general, the more independent claims required, the broader the categories of technologies covered and the higher the technical value they may have [11, 32].

Three main "patent types" are granted by SIPO; these are invention, utility model, and design patents. In the Patent Law of PRC, invention patent is "any new technical solution relating to a product, a process, or an improvement thereof." Utility model patent is "any new technical solution relating to a product's shape, structure, or a combination thereof, which is fit for practical use." Design patent is "any new design of a product's shape, pattern, or a combination thereof, as well as its combination with the color and the shape or pattern of a product, which creates an aesthetic feeling and is fit for industrial application" [33–34].

"Documentation code" is the "standard code for the identification of different kinds of patent documents" and is also called the WIPO Standard ST.16 code [35], which is a letter or a combination of a letter and a number used to distinguish the kind of patent document. This code can reflect the patent type and status information.

"Patent family" is "a set of patents taken in various countries to protect a single invention (when the first application in a country – the priority – is then extended to other offices)" [36]. "Simple patent family" is a group of patents where all patent documents have exactly identical priority date or combination of priority dates. If patent technology is granted globally, then the buyer will be willing to pay a high price. Hence, a high count of simple family members contains high economic and legal value. To combine the concept of patent family, we studied "the count of family citing times." If the count of family citing times of a patent is high, then the patent lays sufficient foundation for future technical improvement and has high economic and technical value [11, 37].

The indicator system of patent value assessment proposed in this work effectively confirmed the results of existing research. Zhang [11], Tsao [14], and Baron et al. [38] stated that number of claims is an important indicator for patent value assessment. Hall [8], Feng [9], Yang [10], and Fischer [38] proposed assessment methods based on patent family and the count of family citing times. In the research of Lanjouw [32], Thoma [40], and Gambardella [41], the assessment indicator system included the number of claims and the count of family citing times. In conclusion, the indicator system of patent value assessment proposed in this work is effective and presents guiding significance for future research.

## V. Conclusion

A wrapper-mode feature selection algorithm was proposed in this work, and multiple UCI standard datasets were used to verify the validity of the algorithm. The experiments demonstrated that the algorithm can effectively reduce the size of the feature set and improve the performance of the classifier. To primarily select indicators for patent value assessment, we applied the feature selection algorithm on data from a patent retrieval database. On the basis of the primary feature selection and to further reduce the size of the indicator system, we employed the SFS algorithm to optimize the indicator system and construct an optimal indicator system that included the number of claims, patent type, documentation code, count of simple family members, and count of family citing times. Finally, an empirical analysis was performed for the indicator system of patent value assessment to fully demonstrate its interpretability and significance.

## Acknowledgments

This work was sponsored by National Natural Science Foundation of China (Grant No. 71804157), the 2018 Science and Technology Program of Fujian Province of China (No. 2018R0095) and the Soft Science Research Program of Fujian Intellectual Property Office (2018ER012).

**References:**
[1] Krattiger, A., Mahoney, R. T., Nelsen, L., Thomson, J. A., Bennett, A. B., Satyanarayana, K., ... & Kowalski, S. (2007). Intellectual Property Management in Health and Agricultural Innovation: A Handbook of Best Practices, Vol. 1.
[2] Yang, X. (2012). Analysis of information retrieval technology patents and research of development trend——take uspto for example. Journal of Intelligence.
[3] Tomo, O. (2011). Ocean Tomo 300 (r) Patent Index. Retrieved January 13, 2011.
[4] China Technology Exchange, M. (2012). Indicators System of Patent Value Analysis, 10-26.
[5] Schankerman, M. (1998). How valuable is patent protection? Estimates by technology field. the RAND Journal of Economics, 77-107.
[6] Lanjouw, J. O., Pakes, A., & Putnam, J. (1998). How to count patents and value intellectual property: The uses of patent renewal and application data. The Journal of Industrial Economics, 46(4), 405-432.
[7] Bessen, J. (2008). The value of US patents by owner and patent characteristics. Research Policy, 37(5), 932-945.
[8] Hall, B. H., Jaffe, A., & Trajtenberg, M. (2005). Market value and patent citations. RAND Journal of economics, 16-38.
[9] Ling, F., Peng, Z., Liu, B., & Che, D. (2015). A latent-citation-network based patent value evaluation method. Journal of Computer Research & Development.
[10] Yang, G. C., Li, G., Li, C. Y., Zhao, Y. H., Zhang, J., & Liu, T., et al. (2015). Using the comprehensive patent citation network (cpc) to evaluate patent value. Scientometrics, 105(3), 1319-1346.
[11] Zhang, Y., Qian, Y., Huang, Y., Guo, Y., Zhang, G., & Lu, J. (2017). An entropy-based indicator system for measuring the potential of patents in technological innovation: rejecting moderation. Scientometrics, 111(3), 1925-1946.
[12] Ercan, S., & Kayakutlu, G. (2014). Patent value analysis using support vector machines. Soft computing, 18(2), 313-328.
[13] Wu, J. L., Chang, P. C., Tsao, C. C., & Fan, C. Y. (2016). A patent quality analysis and classification system using self-organizing maps with support vector machine. Applied Soft Computing, 41, 305-316.
[14] Tsao, C. C., Chang, P. C., Fan, C. Y., Chang, S. H., & Phillips, F. (2017). A patent quality classification model based on an artificial immune system. Soft Computing, 21(11), 2847-2856
[15] Dereli, T., & Durmuşoğlu, A. (2009). Classifying technology patents to identify trends: Applying a fuzzy-based clustering approach in the Turkish textile industry. Technology in Society, 31(3), 263-272.
[16] Dereli, T., Baykasoğlu, A., Durmuşoğlu, A., & Durmuşoğlu, Z. D. (2011). Enhancing technology clustering through heuristics by using patent counts. Expert Systems with Applications, 38(12), 15383-15391.
[17] Breiman, L., Friedman, J., Stone, C. J., & Olshen, R. A. (1984). Classification and regression trees. CRC press.

[18] Cortes, C., & Vapnik, V. (1995). Support-vector networks. Machine learning, 20(3), 273-297.

[19] Altman, N. S. (1992). An introduction to kernel and nearest-neighbor nonparametric regression. The American Statistician, 46(3), 175-185.

[20] Aha, D. W., & Bankert, R. L. (1996). A comparative evaluation of sequential feature selection algorithms. In Learning from Data (pp. 199-206). Springer New York.

[21] Chandrashekar, G., & Sahin, F. (2014). A survey on feature selection methods. Computers & Electrical Engineering, 40(1), 16-28

[22] Guyon, I., & Elisseeff, A. (2003). An introduction to variable and feature selection. Journal of machine learning research, 3(Mar), 1157-1182.

[23] Blake, C. L. (1998). UCI repository of machine learning databases. http://www. ics. uci. edu/~ mlearn/MLRepository. html.

[24] Yihui, Q., & Chiyu, Z. (2016). Research of indicator system in customer churn prediction for telecom industry. In Computer Science & Education (ICCSE), 2016 11th International Conference on (pp. 123-130). IEEE.

[25] Luan, L., & Genlin, J. I. (2004). The study on decision tree classification techniques. Computer Engineering, 30(9), 94-95.

[26] Priyama, A., Abhijeeta, R. G., Ratheeb, A., & Srivastavab, S. (2013). Comparative Analysis of Decision Tree Classification Algorithms. International Journal of Current Engineering and Technology, 3(2), 334-337.

[27] Quinlan, J. R. (1986). Induction of decision trees. Machine learning, 1(1), 81-106.

[28] Quinlan, J. R. (1996). Improved use of continuous attributes in C4. 5. Journal of artificial intelligence research, 4, 77-90.

[29] Mehta, M., Agrawal, R., & Rissanen, J. (1996). SLIQ: A fast scalable classifier for data mining. Advances in Database Technology—EDBT'96, 18-32.

[30] Breiman, L., Friedman, J., Stone, C. J., & Olshen, R. A. (1984). Classification and regression trees. CRC press.

[31] Beijing Incoshare CO., LTD. (2011). Technical Innovative Intelligence Platform. http://www.incopat.com/

[32] Lanjouw, J. O., & Schankerman, M. (1997). Stylized facts of patent litigation: Value, scope and ownership (No. w6297). National Bureau of Economic Research.

[33] Wang, M. H., Hsiao, Y. C., Tsai, B. H., Lee, C. S., & Lin, T. T. (2015, May). Fuzzy markup language with genetic learning mechanism for invention patent quality evaluation. In Evolutionary Computation (CEC), 2015 IEEE Congress on (pp. 251-258). IEEE.

[34] Sidel, M. (1985). Copyright, Trademark and Patent Law in the People's Republic of China. Tex. Int'l LJ, 21, 259.

[35] World Intellectual Patent Organization. (2009). Handbook on industrial property information and documentation. Geneva: World Intellectual Patent Organization (WIPO).

[36] Publishing, O. (2001). OECD Science, Technology and Industry Scoreboard 2001 Towards a Knowledge-based Economy. Organisation for Economic Co-operation and Development.

[37] Harhoff, D., Scherer, F. M., & Vopel, K. (2003). Citations, family size, opposition and the value of patent rights. Research policy, 32(8), 1343-1363.

[38] Baron, J., & Delcamp, H. (2010). Patent quality and value in discrete and cumulative innovation.

[39] Fischer, T., & Leidinger, J. (2014). Testing patent value indicators on directly observed patent value—An empirical analysis of Ocean Tomo patent auctions. Research Policy, 43(3), 519-529. [40]

[40] Thoma, G. (2014). Composite value index of patent indicators: Factor analysis combining bibliographic and survey datasets. World patent information, 38, 19-26.

[41] Gambardella, A., Harhoff, D., & Verspagen, B. (2008). The value of European patents. European Management Review, 5(2), 69-84.